\documentclass[letterpaper]{article}
\usepackage{aaai}
\usepackage{times}
\usepackage{algorithm}
\usepackage{algpseudocode}
\usepackage{helvet}
\usepackage{courier}
\usepackage{float} 
\usepackage{amsmath,amssymb}
\usepackage{caption}
\usepackage{graphicx}
\usepackage{multirow}
\usepackage{amsmath}
\DeclareMathOperator*{\argmax}{arg\,max}
\DeclareMathOperator{\E}{\mathbb{E}}
 
\begin{document}
%
\title{Teaching a Robot to Walk \\Using Reinforcement Learning}
\author{Jack Dibachi\ and Jacob Azoulay\\
Stanford University\\
AA228: Decision Making under Uncertainty\\
dibachi@stanford.edu  |  jazoulay@stanford.edu\\
}
\maketitle
\begin{abstract}
\begin{quote}
Classical control techniques such as PID and LQR have been used effectively in maintaining a system state, but these techniques become more difficult to implement when the model dynamics increase in complexity and sensitivity. For adaptive robotic locomotion tasks with several degrees of freedom, this task becomes infeasible with classical control techniques. Instead, reinforcement learning can train optimal walking policies with ease. We apply deep Q-learning and augmented random search (ARS) to teach a simulated two-dimensional bipedal robot how to walk using the OpenAI Gym \verb!BipedalWalker-v3! environment. Deep Q-learning did not yield a high reward policy, often prematurely converging to suboptimal local maxima likely due to the coarsely discretized action space. ARS, however, resulted in a better trained robot, and produced an optimal policy which officially ``solves" the \verb!BipedalWalker-v3! problem. Various naive policies---including a random policy, a manually encoded inch forward policy, and a stay still policy---were used as benchmarks to evaluate the proficiency of the learning algorithm results. 
\end{quote}
\end{abstract}

\section{Problem Statement} 
Creating feedback controllers capable of maintaining stability in a bipedal robot is a complicated task, particularly because of model complexity and sensitivity. Even when controllers can be made, they are extremely sensitive to variations in the environment, such as irregularities in the walking surface or unexpected obstacles \cite{morimoto}. Moreover, mitigating state uncertainty by including range sensors and contact sensors requires the control designer to account for any number of possible environments. In short, creating a multi-input-multi-output feedback controller to address this control task is intractable.

Instead, the robot's dynamics can be modeled and tested in numerous simulation trials, and use reinforcement learning to determine the optimal policy mapping the robot's state to the action required to walk forward. In the OpenAI Gym's \verb!BipedalWalker-v3! environment, the reward system is defined by incrementally gaining 300 points if the finish line is reached, losing 100 points if robot falls over, and gradually losing points based on the amount of control effort exerted from the motors. Solving the problem is defined by collecting over 300 reward points for 100 consecutive episodes. We use OpenAI Gym's \verb!BipedalWalker-v3! for the robot dynamics and environment, and an augmented random search (ARS) strategy for determining the optimal policy.

\section{Other Work} 
Robot locomotion has been treated as a learning task more frequently in the past few years. \citeauthor{morimoto} apply a model-based approach to a simple five-link bipedal robot using Poincare maps. The policy, however, is only determined using the robot's intrinsic state, such as the joint angles, positions, and velocities, rather than including measurements such as foot contacts and range measurements. This approach, then, cannot be used in environments with surface irregularities (\verb!BipedalWalker-v3!) or obstacles (\verb!BipedalWalkerHardcore-v3!). 
\citeauthor{li} effectively implement reinforcement learning to teach the Cassie robot to walk, though their approach uses a gait library as input to a deep reinforcement learning network \cite{li}. While effective, their objective is to enable robust online planning for environmental uncertainty on physical robotic platforms. 
\citeauthor{mania} outline an ARS strategy, validated in a number of MuJoCo locomotion environments via OpenAI Gym with promising results. ARS is a particularly useful learning strategy for locomotion tasks, especially \verb!BipedalWalker-v3!, as the policy space is filled with locally optimal locomotion strategies \cite{mania}. Other model-free learning strategies require many more episodes to solve these kinds of locomotion tasks, if they solve them at all. For this reason, we use ARS to train the bipedal walker.

\section{Approach}
\begin{figure}
    \centering
    \includegraphics[width=0.45\textwidth]{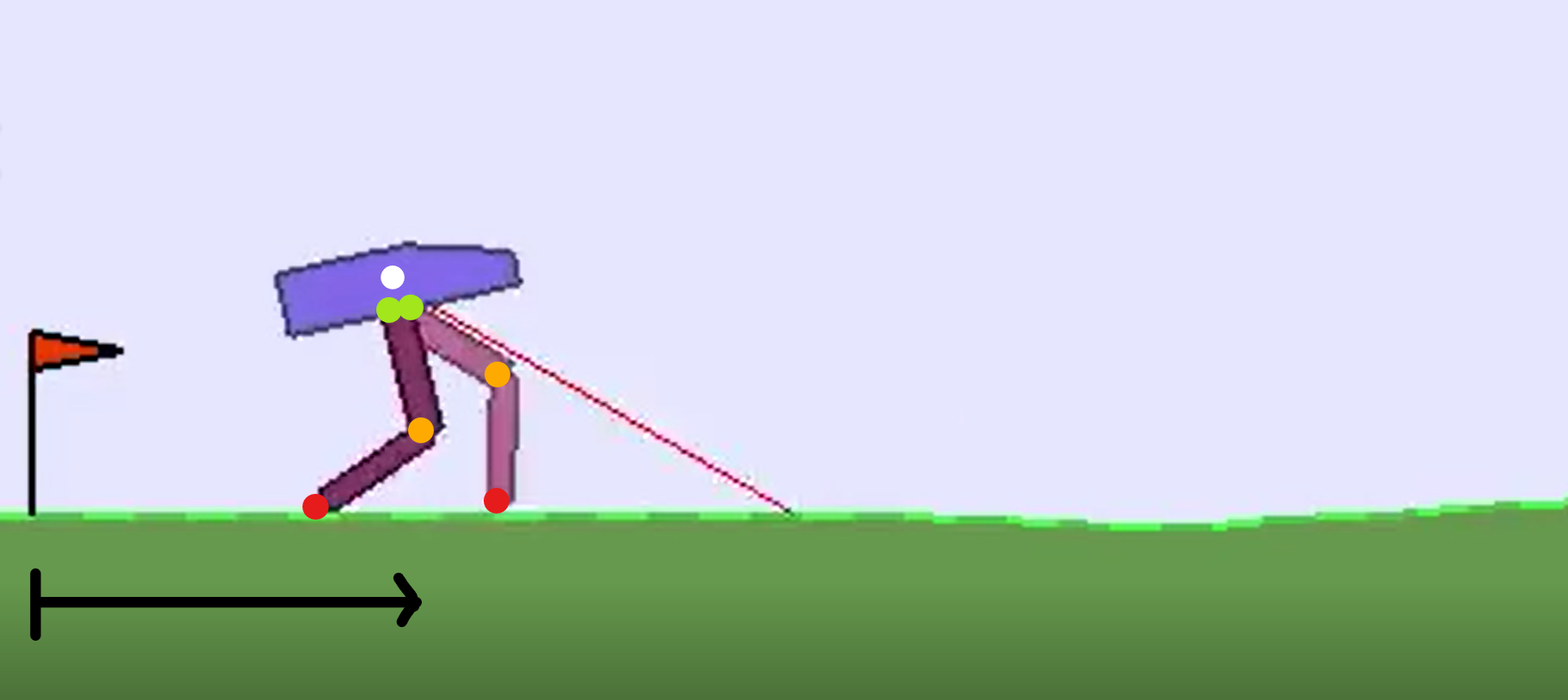}
    \caption{The robot in the \texttt{BipedalWalker-v3} environment. Red flag indicates starting position.}
    \label{fig:robot_env}
\end{figure}

OpenAI Gym's \verb!BipedalWalker-v3! environment provides a model of a five-link bipedal robot, depicted in Figure \ref{fig:robot_env}. The robot state is a vector with 24 elements: $\theta, \dot{x}, \dot{y}, \omega$ of the hull center of mass (white), $\theta, \omega$ of each joint (two green, two orange), contacts with the ground (red), and 10 lidar scans of the ground (red line). The action space, then, is a four element vector representing the torque command to the motor at each of the four joints \cite{gym}.
This robot has to navigate through the environment, which is a randomly seeded ground with an uneven surface. The lidar scans can help detect this unevenness, though they are most useful in the hardcore version of the environment where obstacles and stairs are present. To find the optimal policy, which maps the robot states to the robot actions yielding the highest reward, we try both deep Q-learning and ARS.

\subsection{Deep Q-Learning}
Traditional Q-learning techniques incrementally estimate the action value function $Q(s, a)$, using information at each step to update the Q-function. This populates a lookup table containing action values for every possible state-action pair. This works well for Markov decision process (MDP) problems with discrete state and action spaces, however is insufficient for problems with large continuous state and action spaces. Fortunately, the Q-learning algorithm can be adapted to problems with continuous spaces by using a parametric approximation of the action value function $Q_\theta(s, a)$ such as a neural network with the weights and biases of the network serving as the function hyperparameters. The input layer consists of one node for each state variable and the output layer consists of nodes representing utilities for each discretized action.

Deep Q-learning relies on the minimization of the loss between the approximate action value function $Q_\theta(s, a)$ and the true optimal action value function $Q^{*}(s, a)$ denoted as 
$$l(\theta)=\frac{1}{2} \E_{(s,a)\sim\pi^{*}}[(Q^{*}(s, a) - Q_\theta(s, a))^2]$$

Applying gradient descent to minimize this loss and using samples to approximate the expectation results in the update rule

$$\theta \leftarrow \theta + \alpha(Q^{*}(s, a) - Q_\theta(s, a))\nabla_\theta Q_\theta (s,a)$$

where $\alpha$ is the learning rate which determines the size of each update step. Lastly, because the true optimal policy $Q^{*}(s, a)$ is unknown, it can be approximated, yielding the finalized update rule:

$$\theta \leftarrow \theta + \alpha(r + \gamma \max_{a'}Q_\theta(s', a') - Q_\theta(s, a))\nabla_\theta Q_\theta (s,a)$$

To implement this algorithm for the bipedal walker, the continuous action space is first discretized into a predetermined number of bins. The continuous action space consists of four action variables (one for each joint), each taking a value between $-1$ and $1$. These were discretized and constrained to only take one of three values $-1, 0,$ and $1$, which results in $3^4 = 81$ possible actions represented as a vector of four numbers. 

\begin{figure}
    \centering
    \includegraphics[width=0.45\textwidth]{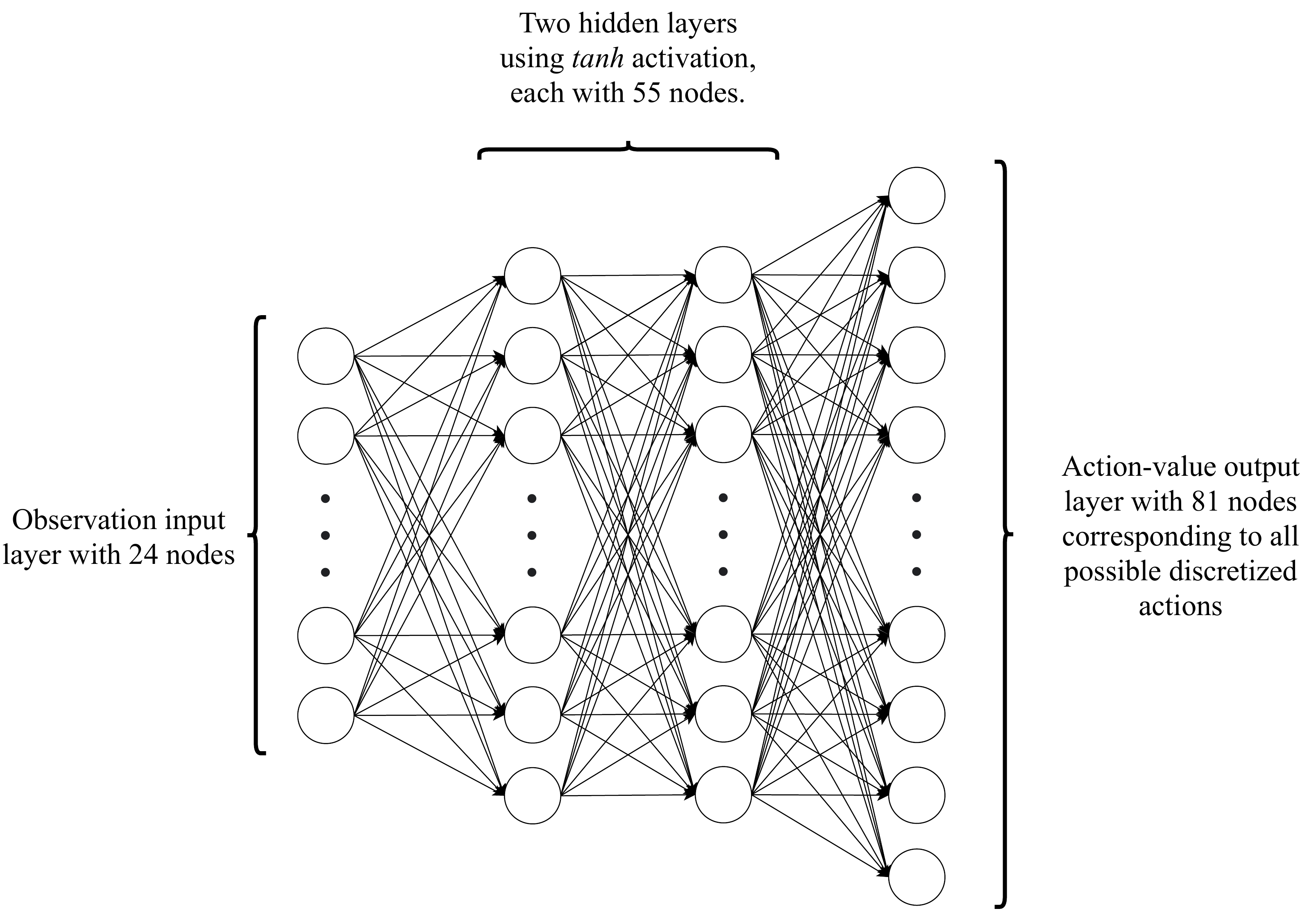}
    \caption{Q-learning neural network structure.}
    \label{fig:Q_NN}
\end{figure}

The neural network representation has 24 input nodes, each representing one of the 24 state variables observed by the agent. The output layer has 81 nodes (one per possible action) which correspond to the approximated utilities of taking an action given the observed state. The output layer uses a linear activation function in order to produce unbounded linear utility outputs. There are two hidden layers, each with 55 nodes using $tanh$ activation functions. A visual representation of the neural network structure is depicted in Figure \ref{fig:Q_NN}.

\begin{algorithm}
\caption{Deep Q-Learning Algorithm}\label{alg:DQL_code}
\begin{algorithmic}
\Ensure Initialize hyperparameters: $\alpha$, $\gamma$, action bin size
    \medskip
    \State Initialize neural net Q with weights $\theta$
    \medskip
    \For {batch in training duration}
        \State $\epsilon \gets$ decay($\epsilon$)
        \medskip
        \State $Q\_ \gets Q$
        \medskip
        \For {episode in batch size}
            \State {$s\gets$ reset environment}
            \medskip
            \While{episode is not $done$}
                \State $a = \begin{cases} \argmax_{a'} Q(s, a') , & \mbox{with prob } 1 -\epsilon \\ \mbox{random action}, & \mbox{with prob } \epsilon \end{cases}$
                \medskip
                \State $s', r$ = observed new state, reward
                \medskip
                \State $Q_{target}= r + (1 - done)\gamma\max_{a'} Q\_(s', a')$
                \medskip
                \State $Q_{output}= Q(s, a)$
                \medskip
                \State $\theta$ updated with gradient descent:
                \medskip
                \State $d\theta \gets d\theta + \frac{\partial (Q\_target - Q\_output)^2}{\partial \theta}$
                \medskip
                \State $s \gets s'$
                \medskip
            \EndWhile
        \EndFor
    \EndFor
\end{algorithmic}
\end{algorithm}

To find the optimal walking policy, we encourage exploration of the action space using $\epsilon$-greedy exploration, where the robot takes some random action with probability $\epsilon$. The amount of exploration needed decreases as the robot improves its gait, so we decay the $\epsilon$ term over time. Furthermore, to minimize unstable learning, training episodes are run in batches, with each batch using a static copy of the current estimation of the action value function (represented as a neural network) when computing $r + \gamma \max_{a'}Q_\theta(s', a')$.

The neural network representations and gradient descent loss minimization are implemented using the Python module \verb!pytorch! \cite{pytorch}.

\subsection{Augmented Random Search (ARS)}
While Q-learning required the discretization of the action space, ARS allows for both a continuous state space and action space, often resulting in efficient optimal policies in fewer training iterations \shortcite{mania}. ARS explores the policy parameter space, as opposed to the action and state space, by taking a set of parameter samples with zero mean Gaussian noise and executing rollouts for each of those samples.

The parameters of the model $\theta$ take the form of an $n{\times}m$ matrix where $n$ is the number of action variables and $m$ is the number of state variables. A state observation is matrix multiplied to produce a desired action to take. In ARS, several parameter noise matrices $\delta$ are sampled, resulting in $\theta + v\delta$ and $\theta - v\delta$ pairs where $\theta$ is the current parameterization, $v$ is a step size scalar that determines how much noise is introduced to the model, and $\delta$ is a random noise matrix of the same dimensions as $\theta$. If in total there are $h$ number of random noise matrices $\delta$ then there are $2h$ matrices, each of which are used to conduct rollouts. These noisy parameterizations all lie within a hypersphere of a specified radius around the current parameterization $\theta$ \shortcite{mania}.

Rewards for each rollout are collected, and their standard deviation $\sigma$ is computed and used to normalize the final update step. The $h$ $v\delta$ matrices are then sorted in descending order of maximum reward using the criteria $\max(r_{\theta + v\delta_i}, r_{\theta - v\delta_i})$ for each parameterization pair $i$, where $r_{\theta + v\delta_i}$ denotes the reward received from the rollout using a model parameterization $\theta + v\delta_i$, and $r_{\theta - v\delta_i}$ denotes the reward received from the rollout using a model parameterization $\theta - v\delta_i$. The top $m$ pairs are then used to update $\theta$ using the following update rule:

$$\theta \leftarrow \theta + \frac{\alpha}{m\sigma}\sum_{i=1}^{m} [(r_{\theta + v\delta_i} - r_{\theta - v\delta_i}) * \delta_i] $$

where $\alpha$ represents the learning rate. 

\begin{figure}[H]
    \centering
    \includegraphics[width=0.45\textwidth]{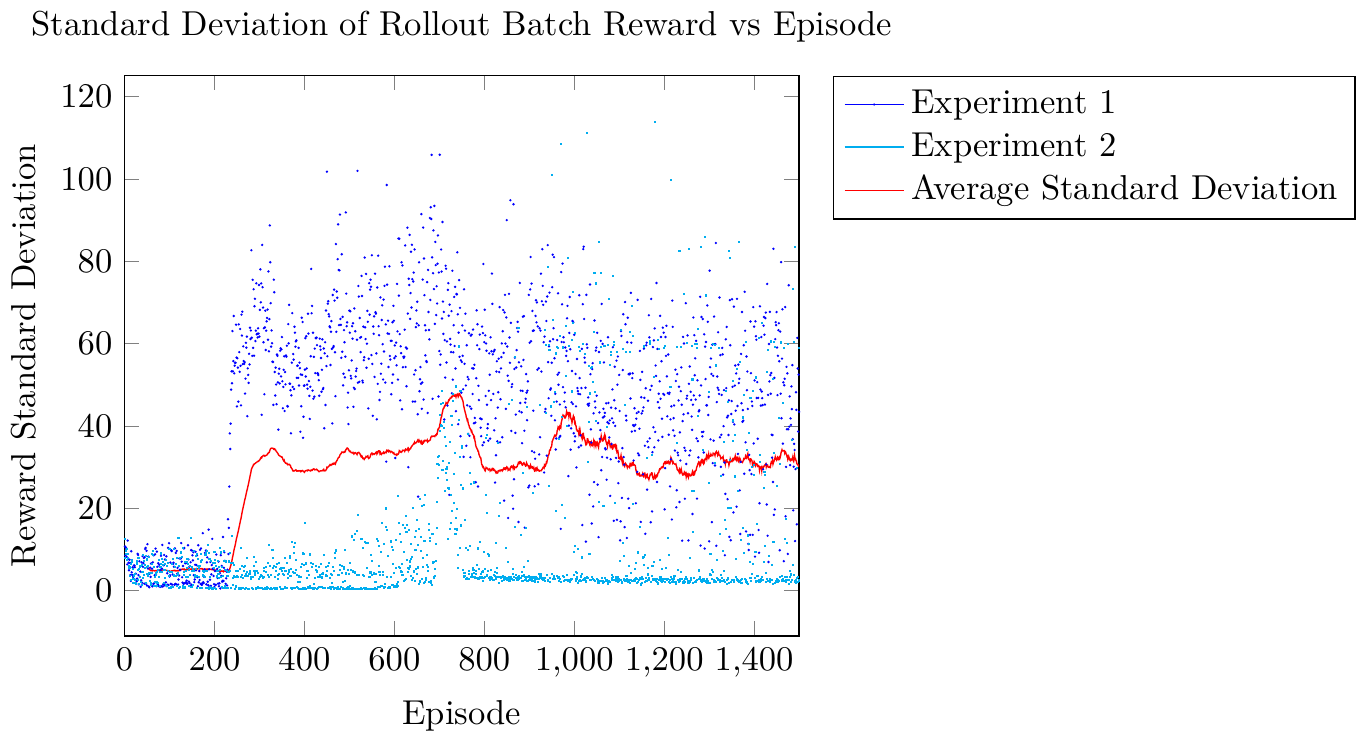}
    \caption{ARS reward standard deviation $\sigma$ over 1500 episodes.}
    \label{fig:alpha002_std}
\end{figure}

As the agent learns and the average reward size increases, the summation over $r_{\theta + v\delta_i} - r_{\theta - v\delta_i}$ will increase as well, resulting in proportionately larger step sizes as the policy improves overtime. To mitigate this inconsistency in step size, the learning rate $\alpha$ is normalized by the standard deviation of the rollout rewards $\sigma$. This normalization is the primary difference between basic random search (BRS) and augmented random search (ARS). The reward standard deviation $\sigma$ overtime is plotted in Figure \ref{fig:alpha002_std}.

Lastly, each state observation is normalized before it is fed through the model to determine an optimal action, eliminating unintentional bias towards state variables with larger absolute bounds.

\begin{algorithm}
\caption{ARS Algorithm}\label{alg:ARS_Code}
\begin{algorithmic}
\Ensure Initialize hyperparameters: $\alpha$, $\theta$, number of noise matrices $h$, noise scalar $v$, number of best performing noise matrices $m$ 
    \medskip
    \For {episode in training duration}
    \State $\delta = [\delta_1:\delta_h]$ where each $\delta_i$ is a randomly sampled 
        \State zero mean noise sample matrix the size of $\theta$
        \medskip
        \For {each $\delta_i$ in $\delta$}
            \State $r_{\theta + v\delta_i} = $ reward collected from rollout using 
            \State $\theta + v\delta_i$ to determine action $a$ at each step 
            \State observation $s$
            \medskip
            \State $r_{\theta - v\delta_i} = $ reward collected from rollout using 
            \State $\theta - v\delta_i$ to determine action $a$ at each step 
            \State observation $s$
            \medskip
            \State $r\_list \gets$ store $(r_{\theta + v\delta_i}, r_{\theta - v\delta_i}, v\delta_i)$ in list 
            \medskip
        \EndFor
        \medskip
        \State $\sigma =$ std of $\{r_{\theta + v\delta_1}, r_{\theta - v\delta_1}, ...,r_{\theta + v\delta_h}, r_{\theta - v\delta_h}\}$
        \medskip
        \State sort $r\_list$ using criterion $\max(r_{\theta + v\delta_i}, r_{\theta - v\delta_i})$
        \medskip
        \State discard $h - m$ worst performing tuples with lowest 
        \State rewards
        \medskip
        \State $\theta \leftarrow \theta + \frac{\alpha}{m\sigma}\sum_{i=1}^{m} [(r_{\theta + v\delta_i} - r_{\theta - v\delta_i}) * \delta_i] $
        \medskip
    \EndFor
\end{algorithmic}
\end{algorithm}

\section{Analysis}
The primary metric indicating how well the agent learns is the change in the average collected reward per episode. The ultimate goal is to achieve high reward collecting performance in as few training iterations as possible. The performance of each algorithm is also evaluated compared to three naive approaches: a random policy model, a manually encoded policy which consists of a set of actions that inch the walker forward, and a policy that has the walker take one step forward and stay still for the duration of the episode iteration.

With both Q-learning and ARS, there are many hyperparameters that can be adjusted when training the bipedal robot how to walk. For instance, in the above implementation of Q-learning the following need to be specified: learning rate $\alpha$, exploration parameter $\epsilon$, discount factor $\gamma$, action space discretization bin sizes, neural network structure (number and size of hidden layers), layer activation functions, batch size, and total number of episode iterations. With ARS, the following hyperparameters must be specified: learning rate $\alpha$, noise scalar $v$, the number of random noise matrices $h$, the number of highest ranked matrices to use $m$, and the total number of episode iterations. It is infeasible to comprehensively explore all combinations of hyperparameters. Through experimental trial runs, however, adjusting the learning rate proved to result in large improvements in learning.

\section{Results} 
\begin{figure}[H]
    \centering
    \includegraphics[width=0.45\textwidth]{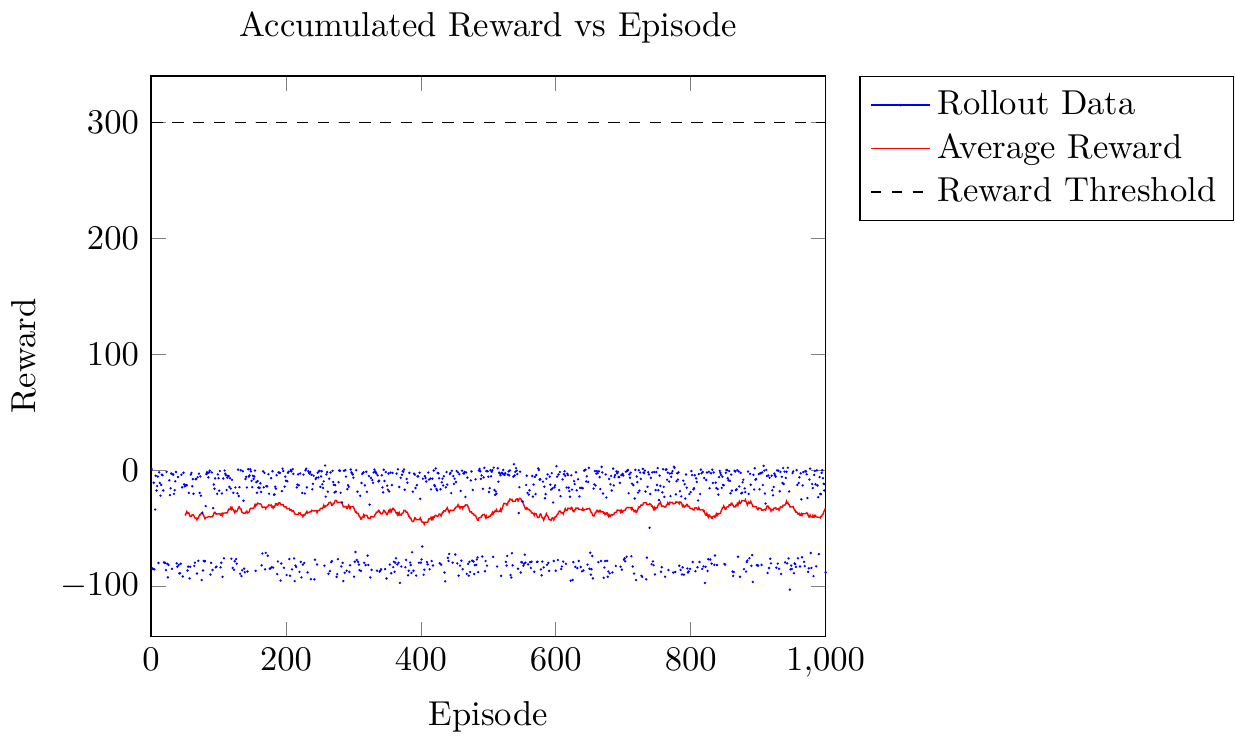}
    \caption{Random policy rolled out for 1000 episodes.}
    \label{fig:random}
\end{figure}

\begin{figure}
    \centering
    \includegraphics[width=0.45\textwidth]{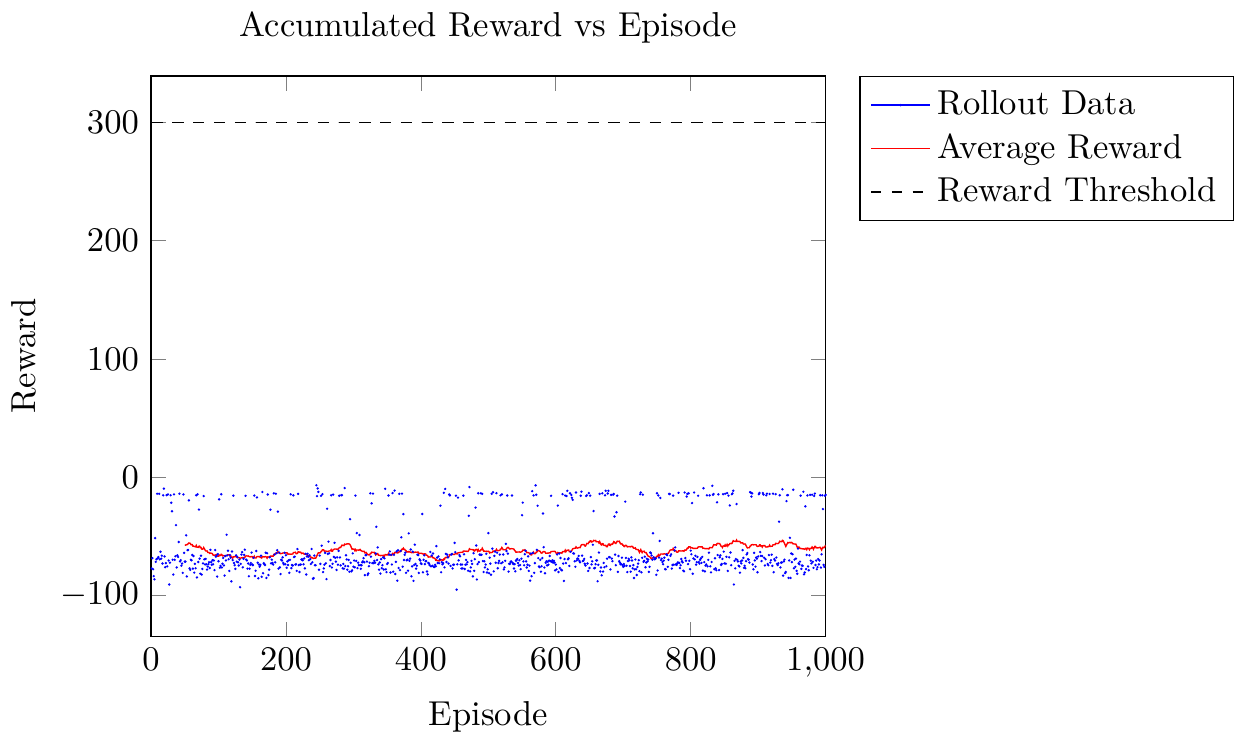}
    \caption{Predetermined crawling agent rolled out for 1000 episodes.}
    \label{fig:predetermined}
\end{figure}

As benchmarks for the ARS performance, we compare accumulated rewards per episode with other agents. The random agent selects a random sample along a uniform distribution from the action space, regardless of the state. Evaluating the random agent for 1000 episodes gives clusters of accumulated reward just below 0 and just above -100 in Figure \ref{fig:random}. Thus, the agent either falls over after some effort forward, or gradually loses points by twitching about, but stays upright. The random agent rarely accumulates a net positive reward in any individual episode, let alone the success threshold of 300 reward points.
However, the random agent still outperforms the predetermined agent shown in Figure \ref{fig:predetermined}, which periodically repeats an inching forward action regardless of state. This agent has fewer episodes with zero reward and a higher percentage of episodes with approximately -100 reward, as the periodic agent either falls over or expends excess control effort. Because the periodic agent tends to stand for longer than the random agent, it spends longer expending control effort without making up for it by covering linear distance, thus resulting in a lower average reward than the random agent. 

\begin{figure}
    \centering
    \includegraphics[width=0.45\textwidth]{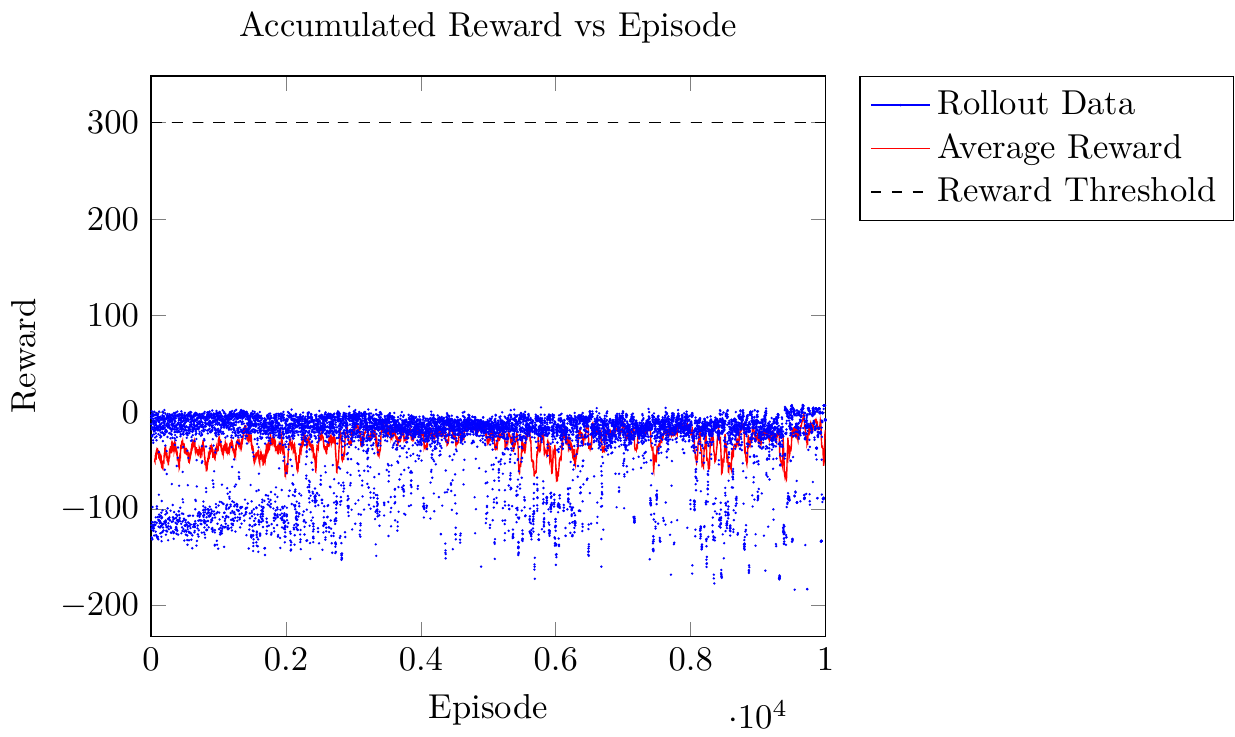}
    \caption{Deep Q-Network performance over 10000 episodes with decaying $\epsilon$-greedy exploration and learning rate $\alpha=0.001$.}
    \label{fig:dqn}
\end{figure}

A more advanced benchmark is the deep Q-learning agent with decaying $\epsilon$-greedy exploration. Although this algorithm takes significantly more episodes (and thus run-time) to obtain results, Figure \ref{fig:dqn} shows the majority of episodes accumulate rewards closer to zero than the random or periodic agents, outperforming both these agents.

\begin{figure}
    \centering
    \includegraphics[width=0.45\textwidth]{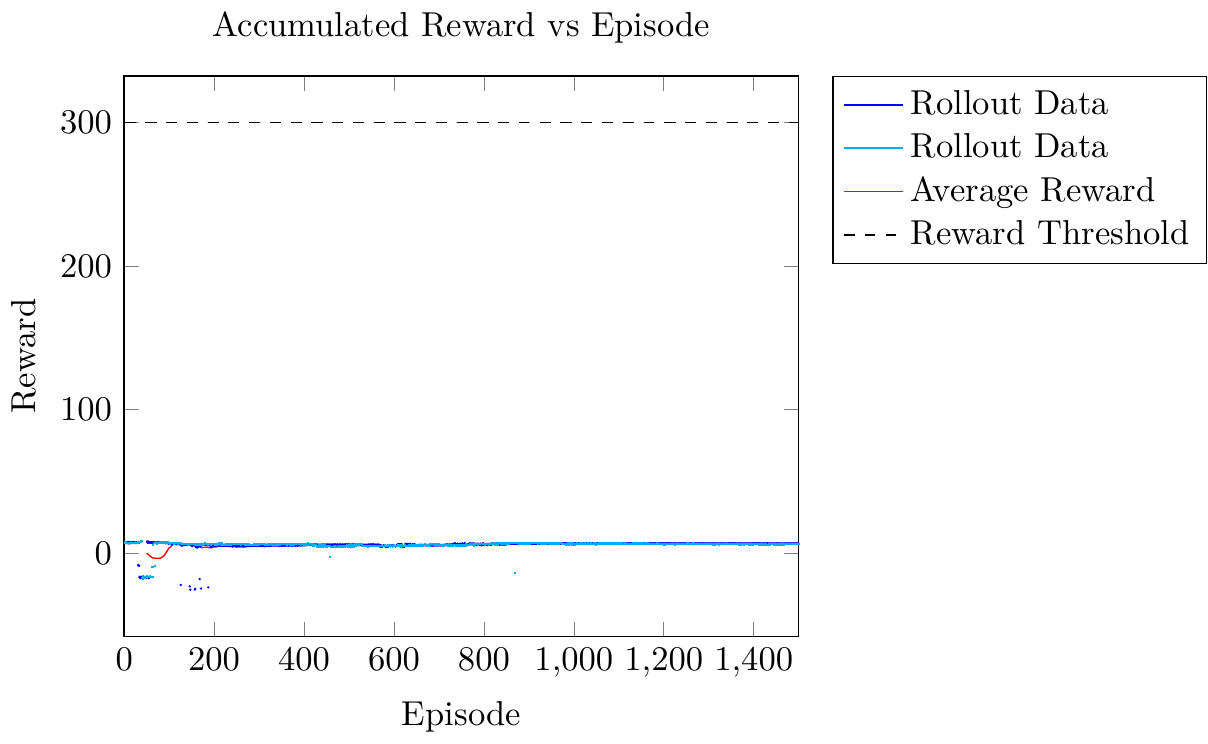}
    \caption{ARS learning curve with $\alpha = 0.001$, averaged over two separate experiments.}
    \label{fig:alpha0001}
\end{figure}

Implementing ARS for the same learning rate gives better performance than the Q-learning approach, in that the average reward is greater than zero, and the robot rarely accumulates rewards less than zero, as shown in two separate experiments in Figure \ref{fig:alpha0001}. This is indicative of the robot taking a single step to stay upright, then staying still to avoid losing reward points. However, there is not enough exploration in the policy space for the agent to take sequential steps. Aggressively increasing the learning rate should remedy this issue.

\begin{figure}
    \centering
    \includegraphics[width=0.45\textwidth]{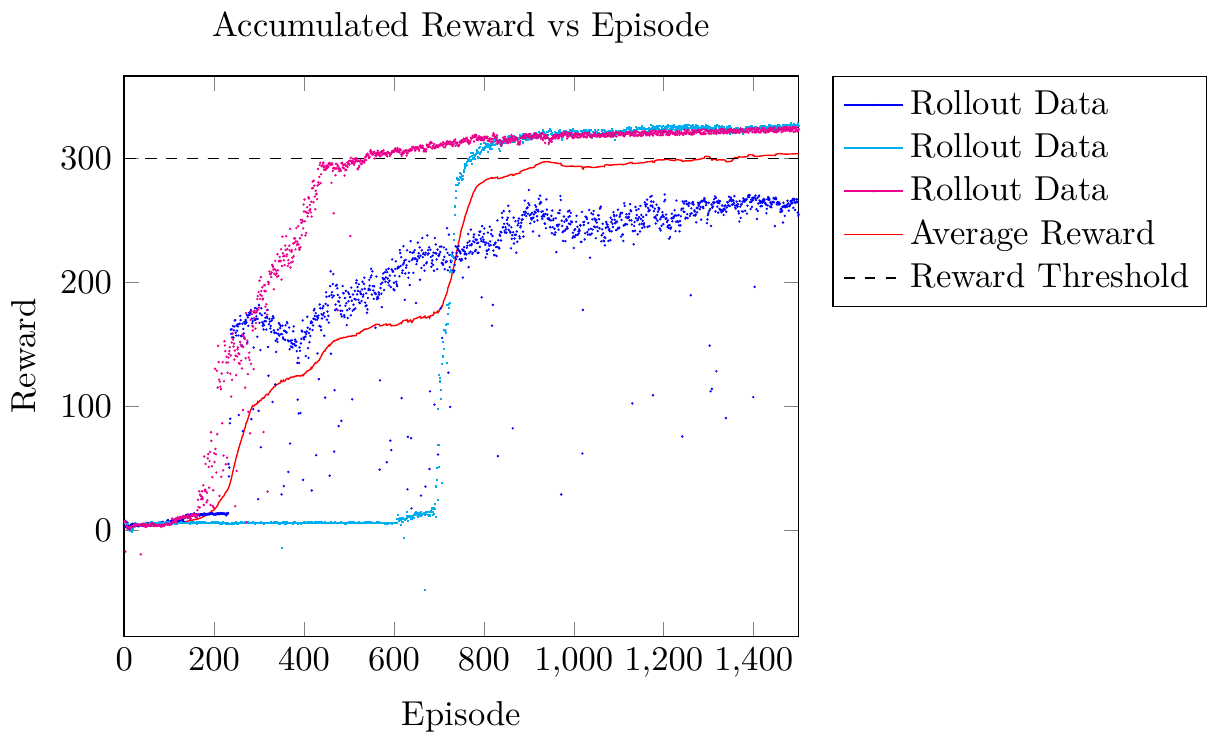}
    \caption{ARS learning curve with $\alpha = 0.02$, averaged over three separate experiments.}
    \label{fig:alpha002}
\end{figure}

The ARS agent with a learning rate of $\alpha=0.02$ demonstrates significantly better performance in accumulated reward than the other agents we have covered so far. Results over three experiments are shown in Figure \ref{fig:alpha002}, illustrating the agent's ability to take multiple steps and maintain balance. The agent's learning curve varies broadly over these experiments because the learning process is stochastic, so the episode in which the agent learns to take a certain stride can happen in a wide range of possible times.  

\begin{figure}[H]
    \centering
    \includegraphics[width=0.45\textwidth]{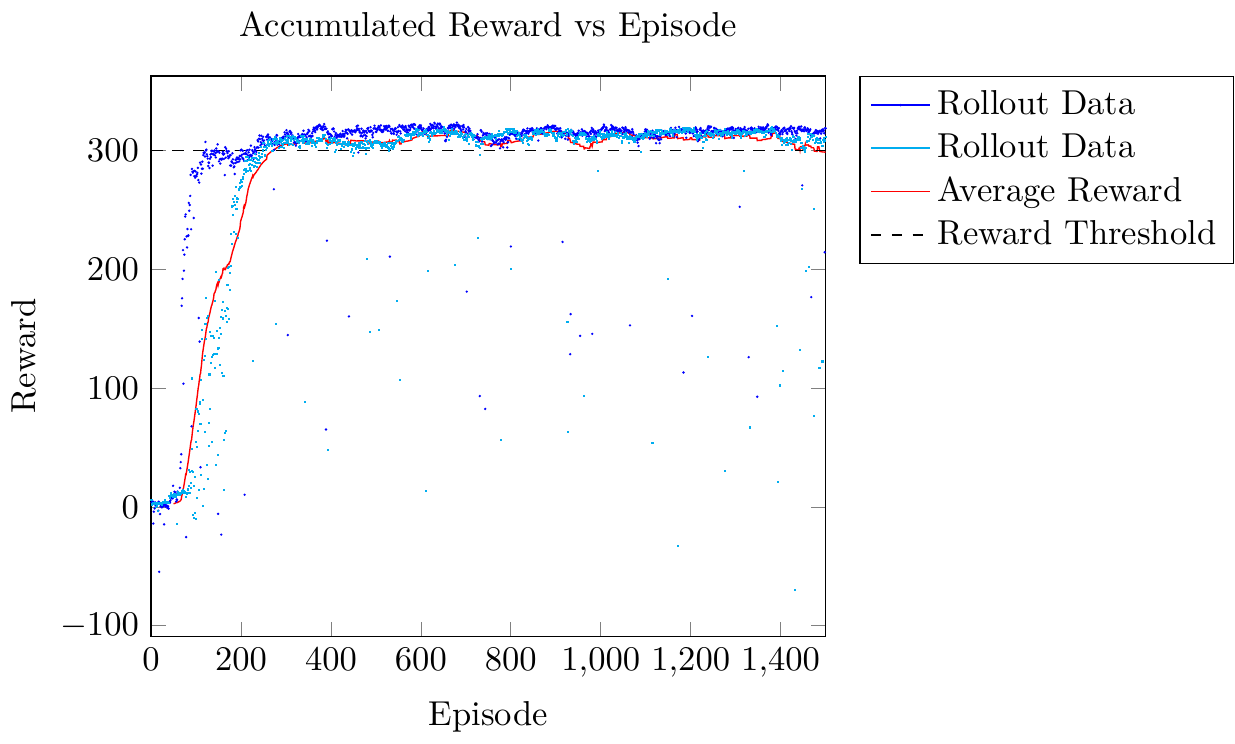}
    \caption{ARS learning curve with $\alpha = 0.06$, averaged over two separate experiments.}
    \label{fig:alpha006}
\end{figure}

Increasing the learning rate to $\alpha=0.06$ demonstrates even higher performance, making sharp increases in reward accumulation in approximately 100 episodes, and clearing the reward threshold in less than 400 episodes. Results from two experiments are shown in Figure \ref{fig:alpha006}.

\begin{tabular}{ |p{0.13\textwidth}||p{0.15\textwidth}|p{0.09\textwidth}|  }
    \hline
    \multicolumn{3}{|c|}{Algorithms and Run Times} \\
    \hline
    Algorithm& Total Run Time (s) & Episodes\\
    \hline
    Random & 181 & 1000\\
    Periodic & 298 & 1000\\
    Stationary & 354 & 1000\\
    Deep Q & 3861 & 10000\\
    ARS $\alpha=0.001$ & 1212 & 1500\\
    ARS $\alpha=0.02$ & 8428 & 1500\\
    ARS $\alpha=0.06$ & 6864 & 1500\\
    \hline
\end{tabular}

\section{Discussion}
When choosing a reinforcement learning method to implement in a system, there is no all-encompassing algorithm that universally produces the best results. It is necessary to not only consider the structure of the problem, but also experiment with various algorithms and various hyperparameters for those algorithms. 

In context of the bipedal walker, the realm of possible algorithms to choose from is narrowed due to the large continuous state space and action space, producing the best results with a model-free method. While with some continuous problems it is possible to discretize the state space and action space, doing so is computationally intractable for the problem at hand since producing a sufficiently granular state space and action space would result in memory overflow as the discretized space scales exponentially with bin size. This motivated the decision to use deep Q-learning and ARS.

The advantage of deep Q-learning lies in the ability to have a continuous state space; however, our implementation still requires discretizing the action-space. The robot did not yield successful learning results likely because of this discretization. Due to the exponential growth in discretized action space with each additional bin, the possible actions the agent could execute were very limited compared to the full continuous action space. Many of the trial runs produced policies which have the robot take one step forward and stay still for the duration of the episode. The robot learned that attempting to move posed a high risk of falling (which is penalized with a large negative reward), and as a result chooses to remain still and receive a higher reward, albeit still negative or close to zero.

Perhaps a more aggressive exploration strategy or a finer action space discretization could nudge the agent out of this local optimum. Furthermore, future work could involve experimenting with different training batch sizes and overall length of training.

ARS, however, yielded consistently reliable learning. According to OpenAI, the \verb!BipedalWalker-v3! is considered ``solved" when the agent obtains an average reward of at least 300 over 100 consecutive episodes. Several runs of ARS in this study achieved this metric, in particular the runs using learning rates $\alpha$ in the range of $[0.01, 0.06]$. One of the benefits of ARS is that it allows for a continuous state space \textit{and} continuous action space, an advantage likely contributing to the remarkable learning.

Future work could involve applying the algorithm to the \verb!BipedalWalkerHardcore-v3! environment, which poses a more challenging problem introducing obstacles and pitfalls. Because the hardcore version of the problem still requires the robot to learn how to walk on flat ground, it may be beneficial to initialize future training models with the output model parameterization $\theta$ of the original non-hardcore version of the problem. Doing so will allow the walker to spend more time exploring the expanded state space imposed by the new obstacles rather than relearning how to walk as well.

\section{Conclusion}
This study presented an application of deep Q-learning and ARS to control a bipedal robot in a simulated environment. While deep Q-learning yielded poor results, ARS demonstrated quick and efficient learning for this application. ARS can be extended beyond a two-dimensional simulated environment, and can ultimately be deployed for real world applications of robot autonomy.

\section{Group Member Contributions} 
Jacob:
\begin{itemize}
    \item Deep Q-Learning code using \verb!pytorch! \shortcite{pytorch}
    \item ARS code \cite{arscode}
    \item Collected raw training data
    \item Wrote the following sections of this paper:
    \subitem Abstract
    \subitem Approach
    \subitem Analysis
    \subitem Discussion
    \subitem Conclusion
\end{itemize}

\noindent Jack:
\begin{itemize}
    \item Deep Q-Learning code using \verb!TensorForce! \shortcite{tensorforce}
    \item ARS code \cite{arscode}
    \item Produced data visualizations using \verb!pgfplots! 
    \item Wrote the following sections of this paper:
    \subitem Problem Statement
    \subitem Other Work
    \subitem Results
    \subitem References
\end{itemize}

\bibliographystyle{aaai}
\bibliography{references.bib}

\end{document}